\documentclass[letterpaper, 10 pt, conference]{ieeeconf}

\IEEEoverridecommandlockouts

\renewcommand{\baselinestretch}{0.99}

\usepackage{import}
\usepackage{wrapfig}

\usepackage{blindtext}
\let\oldbt\blindtext
\renewcommand{\blindtext}[1]{\textcolor{gray}{\oldbt{}}}
\usepackage{times}
\usepackage{bm}

\usepackage[usenames,dvipsnames]{xcolor}

\usepackage{amsmath,amssymb,mathtools}

\usepackage{amsthm}
\usepackage{bbold,dsfont,bbm}
\usepackage{algorithm}
\usepackage[noend]{algpseudocode}

\usepackage{graphicx}
\usepackage{tikz,tikz-cd,pgfplots}
\usepgfplotslibrary{fillbetween}

\usepackage{hyperref}
\usepackage[capitalise,noabbrev]{cleveref}
\usepackage{cite}

\usepackage{multirow}
\usepackage{booktabs}

\usepackage[acronym]{glossaries}

\usepackage{comment}

\usepackage{units}

\usepackage[font=footnotesize]{caption}
\usepackage{subcaption}

\definecolor{baiocchi}{RGB}{193,221,245}

\newcommand*\circled[1]{\tikz[baseline=(char.base)]{
            \node[shape=circle,black,inner sep=1pt, fill=baiocchi] (char) {#1};}}

\newacronym{abk:av}{AV}{autonomous vehicle}
\newcommand{\acc}{\mathrm{ACC}}

\newacronym{abk:bev}{BEV}{battery electric vehicle}

\newacronym{abk:cpo}{CPO}{complete partial order}
\newacronym{abk:cdp}{CDP}{co-design problem}
\newacronym{abk:cdpi}{CDPI}{co-design problem with implementation}

\definecolor{dpred}{rgb}{0.7, 0.0, 0.0}

\newacronym{abk:dp}{DP}{design problem}
\newacronym{abk:dpi}{MDPI}{Monotone design problem with implementation}
\newacronym{abk:dcpo}{DCPO}{directed complete partial order}

\definecolor{dpgreen}{rgb}{0.0, 0.5, 0.0}
\newcommand{\F}[1]{{\color{dpgreen}#1}}
\newcommand{\Ftext}[1]{\textcolor{dpgreen}{#1}}

\newcommand{\fn}{\mathsf{FN}}
\newcommand{\fp}{\mathsf{FP}}

\newcommand{\mat}[1]{\mathbf{#1}}

\newcommand{\op}{^{\mathrm{op}}}

\def\prov{\mathsf{prov}}
\newacronym{abk:poset}{poset}{partially ordered set}

\newcommand{\power}[1]{\mathcal{P}(#1)}

\newcommand{\Rtext}[1]{\textcolor{dpred}{#1}}

\newcommand{\R}[1]{{\color{dpred}#1}}

\def\req{\mathsf{req}}

\newcommand{\setOfFunctionalities}[1]{\F{\mathcal{F}_{#1}}}
\newcommand{\setOfFunctionalitiesOp}[1]{\F{\mathcal{F}_{#1}}^{\mathrm{op}}}
\newcommand{\setOfImplementations}[1]{\mathcal{I}_{#1}}
\newcommand{\setOfResources}[1]{\R{\mathcal{R}_{#1}}}

\newcommand{\trace}[1]{\mathsf{Tr}\left(#1\right)}

\newcommand{\tup}[1]{\langle#1\rangle}

\newcommand{\D}{\mathrm{d}}

\newtheorem{theorem}{Theorem}

\theoremstyle{definition}

\newtheorem{definition}[theorem]{Definition}
\newtheorem{example}[theorem]{Example}
\theoremstyle{remark}
\newtheorem*{remark}{Remark}

\makeatletter
    \if@cref@capitalise
        \crefname{subsection}{Section}{Sections}
        \crefname{subsubsection}{Section}{Sections}
        \crefname{assump}{Assumption}{Assumptions}
        \crefname{problem}{Problem}{Problems}
    \else
        \crefname{subsection}{section}{sections}
        \crefname{subsubsection}{section}{sections}
        \crefname{assump}{assumption}{assumptions}
        \crefname{problem}{problem}{problems}
    \fi
\makeatother

\usetikzlibrary{positioning,intersections, hobby, patterns, calc, decorations.pathmorphing, decorations.markings, shadows,shapes, cd,arrows.meta, fit,quotes}

\tikzset{
   tick/.style={postaction={
      decorate,
      decoration={markings, mark=at position 0.5 with {\draw[-] (0,.4ex) -- (0,-.4ex);}}}
   }
}
\tikzstyle{block} = [draw, rectangle, minimum height=2em, minimum width=3em,font=\bfseries,rounded corners,thick]
\tikzstyle{block} = [draw, rectangle, minimum height=2em, minimum width=3em]
\tikzstyle{block1} = [draw, rectangle, minimum height=1.5em, minimum width=2.5em]
\tikzstyle{blockDyn} = [draw, rectangle, minimum height=2.5em, minimum width=3.5em, align=center, inner sep=10pt, thick, fill=white, copy shadow={draw=black,fill=black,opacity=1,shadow xshift=0.5ex,shadow yshift=-0.5ex}]
\tikzstyle{blockAlg} = [draw, rectangle, minimum height=1.5em, minimum width=2.5em, align=center, inner sep=10pt, thick]
\tikzstyle{sum} = [draw,circle]

\tikzstyle{nodePre} = [circle, draw,inner sep=1pt,node contents={$\preceq$},thick]
\tikzstyle{nodePreEmpty} = [circle, draw,inner sep=1pt,thick]
\tikzstyle{nodePos} = [circle, draw,inner sep=1pt,node contents={$\posceq$},thick]
\tikzstyle{nodeProd} = [rectangle, draw,inner sep=4pt,node contents={$\times$},rounded corners,thick]
\tikzstyle{nodeSum} = [rectangle, draw,inner sep=4pt,node contents={$\mathbf{+}$},rounded corners,thick]

\definecolor{red}{rgb}{0.75, 0.0, 0.0}

\tikzset{fcname/.store in =\fcname, fcname={}}
\tikzset{funame/.store in =\funame, funame={}}
\tikzset{rcname/.store in =\rcname, rcname={}}
\tikzset{runame/.store in =\runame, runame={}}
\tikzset{whereres/.store in =\whereres, whereres=0.5}
\tikzset{wherefun/.store in =\wherefun, wherefun=0.5}
\tikzset{relres/.store in =\relres, relres={above}}
\tikzset{relfun/.store in =\relfun, relfun={above}}
\tikzset{posres/.store in =\posres, posres=1}
\tikzset{posfun/.store in =\posfun, posfun=1}
\tikzset{loos/.store in =\loos, loos=2}
\tikzset{feedback/.store in =\feedback, feedback=0}
\tikzset{
   DP/.style={%
      label/.style={
         font=\everymath\expandafter{\the\everymath\scriptstyle},
         inner sep=5pt,
         node distance=2pt and -2pt},
      semithick,
      node distance=1 and 1,
      rconn/.style={color=white,opacity=0.0,postaction={decorate}, shorten <=3.2pt, shorten >= 0.8,
      decoration={markings, 
      mark= at position 0 with {
               \coordinate (a);
      },
      mark=at position .5 with
      {
              \ifthenelse{\equal{\feedback}{1}}{\def\angleOut{90}\def\angleIn{90}}{\def\angleOut{0}\def\angleIn{180}}    
              \coordinate (b);
              \draw[dashed,dpred,opacity=1.0] (a) to[out=\angleOut,in=\angleIn,looseness=\loos] 
              node[pos=\posres,\relres=\whereres mm,dpred,opacity=1,fill=white,inner sep=1pt,outer sep=1pt]{\footnotesize{\rcname}} (b);
      },
      mark= at position 1 with 
      {
             \ifthenelse{\equal{\feedback}{1}}{\def\angleOut{0}\def\angleIn{0}}{\def\angleOut{180}\def\angleIn{0}} 
              \ifthenelse{\equal{\feedback}{1}}{\def\symbol{\succeq}}{\def\symbol{\preceq}} 
              \coordinate (c);
              \draw[dpgreen,opacity=1.0] (c) to[out=\angleOut,in=\angleIn,looseness=\loos]
              node[pos=\posfun,\relfun=\wherefun mm,dpgreen,opacity=1,fill=white,inner sep=1pt,outer sep=1pt]{\footnotesize{\fcname}} (b){}; %
              \node[draw,circle,inner sep=0.5pt,color=black,fill=white,opacity=1.0] at (b) (nodepreceq) {$\symbol$}; 
      }
      }},
      runconn/.style={color=dpred,dashed,postaction={decorate},
      decoration={markings,
      mark= at position 1 with {
              \coordinate (a);
              \draw[dpred,opacity=1.0,dashed] ($(a)+(0.05,0)$) --++ (0.5,0) node[\relres,pos=\posres]{\footnotesize{\runame}};}
      }
      },
      funconn/.style={color=white,postaction={decorate},
      decoration={markings,
      mark= at position 0 with {
      \coordinate (a);
      \draw[dpgreen] ($(a)+(-0.05,0)$) -- ($(a)+(-0.5,0)$) node[\relfun, pos=\posfun]{\footnotesize{\funame}};}
      }
      },
      execute at begin picture={\tikzset{
         x=\dpx, y=\dpy,
         every fit/.style={inner xsep=\dpx, inner ysep=\dpy}}}
      },
   dpx/.store in=\dpx,
   dpx = 1.5cm,
   dpy/.store in=\dpy,
   dpy = 1.5ex,
   dp port sep/.store in=\dpportsep,
   dp port sep=2,
   dp port length/.store in=\dpportlen,
   dp port length=4pt,
   dp min width/.store in=\dpminwidth,
   dp min width=0.5cm,
   dp rounded corners/.store in=\dpcorners,
   dp rounded corners=2pt,
   dp small/.style={dp port sep=1, dp port length=2.5pt, dpx=.4cm, dp min width=.4cm, dpy=.7ex},
   dp/.code 2 args={%
      \pgfmathsetlengthmacro{\dpheight}{\dpportsep * (max(#1,#2)) * \dpy}
      \pgfkeysalso{draw,%
        minimum width=\dpminwidth,%
        minimum height=\dpheight,%
        font=\bfseries,
        outer sep=0pt,%
        inner sep=5pt,%
        rounded corners=\dpcorners,
        thick,
        prefix after command={\pgfextra{\let\fixname\tikzlastnode}},
        append after command={\pgfextra{\draw
            \ifnum #1=0{} \else foreach \i in {1,...,#1} { 
            ($(\fixname.north west)!{\i/(#1+1)}!(\fixname.south west)$) +(0,0) node[solid,left,circle,color=dpgreen,draw,fill=dpgreen,scale=0.3] {} coordinate (\fixname_fun\i) -- +(0,0) coordinate (\fixname_fun\i')}\fi %
            \ifnum #2=0{} \else foreach \i in {1,...,#2} {
            ($(\fixname.north east)!{\i/(#2+1)}!(\fixname.south east)$) +(0,0) coordinate (\fixname_res\i') -- +(0,0) node[solid,right,circle,color=dpred,draw,fill=dpred,scale=0.3] {} coordinate (\fixname_res\i)}\fi;
         }}}
         },
      dp name/.style={append after command={\pgfextra{\node[label=center,inner sep=2pt,fill=white] at (\fixname) {\textbf{#1}};}}}
   }

\crefname{equation}{}{}
\crefname{figure}{Fig.}{}
\crefname{definition}{Def.}{}
\usepackage[absolute,overlay]{textpos}

\newboolean{extended}
\setboolean{extended}{false}
\ifthenelse{\boolean{extended}}
{
\title{
\textbf{Co-Design of Embodied Intelligence (Extended Version)}
}
\author{Gioele Zardini$^1$, Dejan Milojevic$^{1,2}$, Andrea Censi$^1$, and Emilio Frazzoli$^1$
\thanks{$^1$Institute for Dynamic Systems and Control,
        ETH Z\"urich, Z\"urich, Switzerland
        {\tt \{gzardini,dejanmi,acensi,efrazzoli\}@ethz.ch}}
\thanks{$^2$Automotive Powertrain Technologies Laboratory, Empa - Swiss Federal Laboratories for Materials Science and Technology, D\"ubendorf, Switzerland}
}}
{
\title{
\textbf{Co-design of Embodied Intelligence: A Structured Approach}
}
\author{Gioele Zardini$^1$, Dejan Milojevic$^{1,2}$, Andrea Censi$^1$, and Emilio Frazzoli$^1$
\thanks{$^1$Institute for Dynamic Systems and Control,
        ETH Z\"urich, Z\"urich, Switzerland
        {\tt \{gzardini,dejanmi,acensi,efrazzoli\}@ethz.ch}}
\thanks{$^2$ Automotive Powertrain Technologies Laboratory, Empa - Swiss Federal Laboratories for Materials Science and Technology, D\"ubendorf, Switzerland.
The first author is supported by the Swiss National Science Foundation under NCCR Automation, grant agreement 51NF40\_180545.}
}}

\begin{document}

\begin{textblock*}{\textwidth}(15mm,18mm) %
\bf \textcolor{NavyBlue}{To appear in the Proceedings of the 2021 IEEE/RSJ International Conference on Intelligent Robots and Systems}
\end{textblock*}

\maketitle
\begin{abstract}
We consider the problem of co-designing embodied intelligence as a whole in a structured way, from hardware components such as propulsion systems and sensors to software modules such as control and perception pipelines. We propose a principled approach to formulate and solve complex embodied intelligence co-design problems, leveraging a monotone co-design theory. The methods we propose are intuitive and integrate heterogeneous engineering disciplines, allowing analytical and simulation-based modeling techniques and enabling interdisciplinarity. We illustrate through a case study how, given a set of desired behaviors, our framework is able to compute Pareto efficient solutions for the entire hardware and software stack of a self-driving vehicle.
\end{abstract}

\ifthenelse{\boolean{extended}}{
\section{Introduction}
\label{sec:introduction}
In the last decade the research on embodied intelligence has observed important developments. While researchers mostly focused on specific problems in robotics, we know very little about the optimal co-design of autonomous robots as a whole. Traditionally, the design of embodied intelligence is approached in a compartmentalized manner, hindering interdisciplinary collaboration and design automation. In particular, current design techniques fail to fill the gap between the specificity of technical results in robotic fields and more general trade-offs regarding energy consumption, computational effort, cost and performance of the autonomous robots; these are two sides of the same coin, and both need to be captured by a comprehensive co-design theory. Some of the unresolved questions are: What is the simplest sensor that a robot can use to obtain a specific performance in a given task? How much computation is really needed? What are the trade-offs between robot safety and task performance? How can one optimally design a robotic platform by minimizing resource usage?

In this paper we present a structured approach to efficiently leverage a monotone theory of co-design to formulate and solve embodied intelligence co-design problems (\cref{fig:dpgeneralembodied}). The approach we propose is intuitive and allows one to co-design robots solving heterogeneous tasks, unifying different modeling techniques, hence promoting interdisciplinarity.

\begin{figure}[t]
    \begin{subfigure}[b]{\linewidth}
    \centering
    \includegraphics[width=0.9\linewidth]{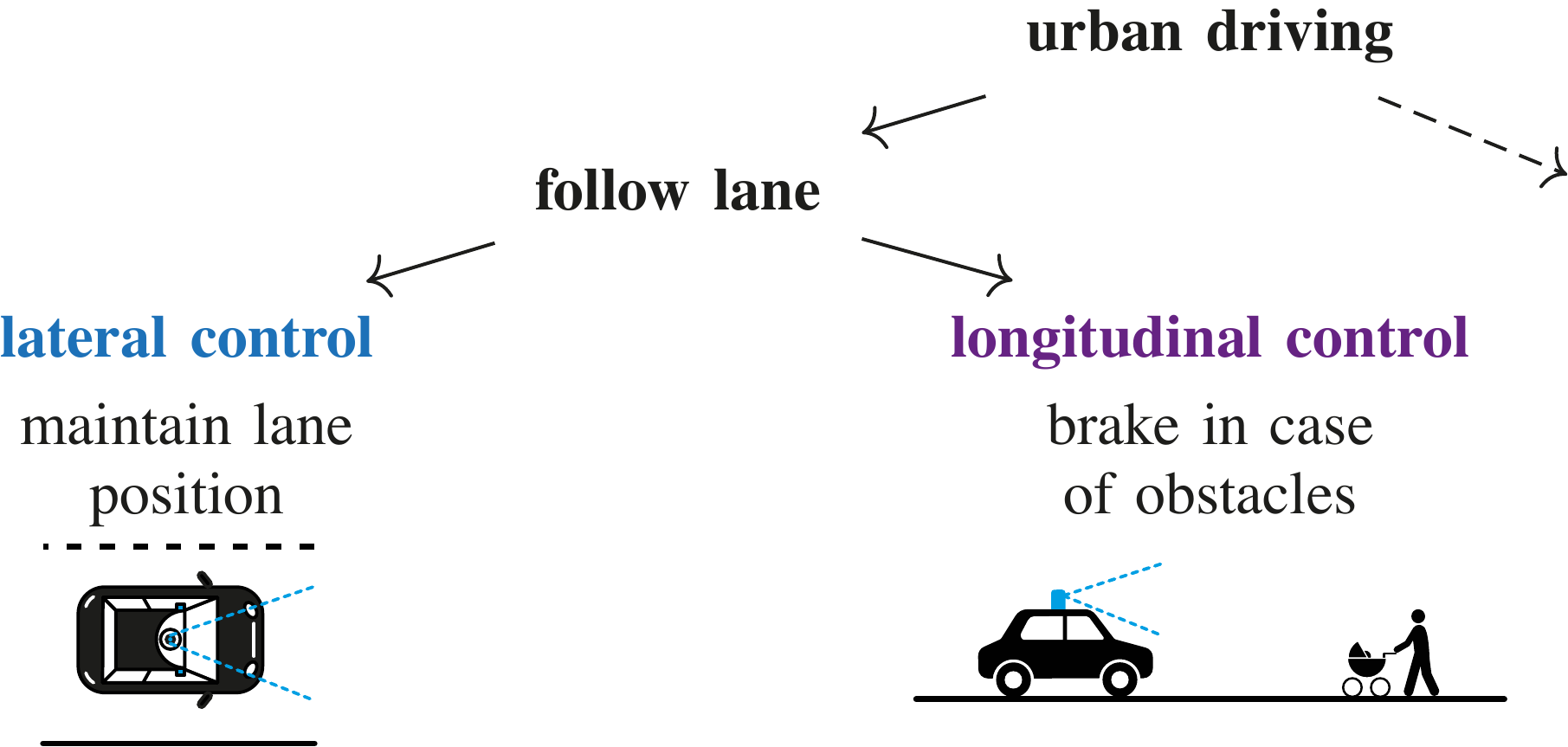}
    \subcaption{We instantiate the functional decomposition approach via the example of the self-driving task for an \gls{abk:av}. Specifically, one can decompose this ability into two more specific functions: lateral and longitudinal control.}
    \label{fig:decomposition}
    \vspace{0.15cm}
    \end{subfigure}
\begin{subfigure}[h]{\columnwidth}
\centering
    \includegraphics[width=\columnwidth]{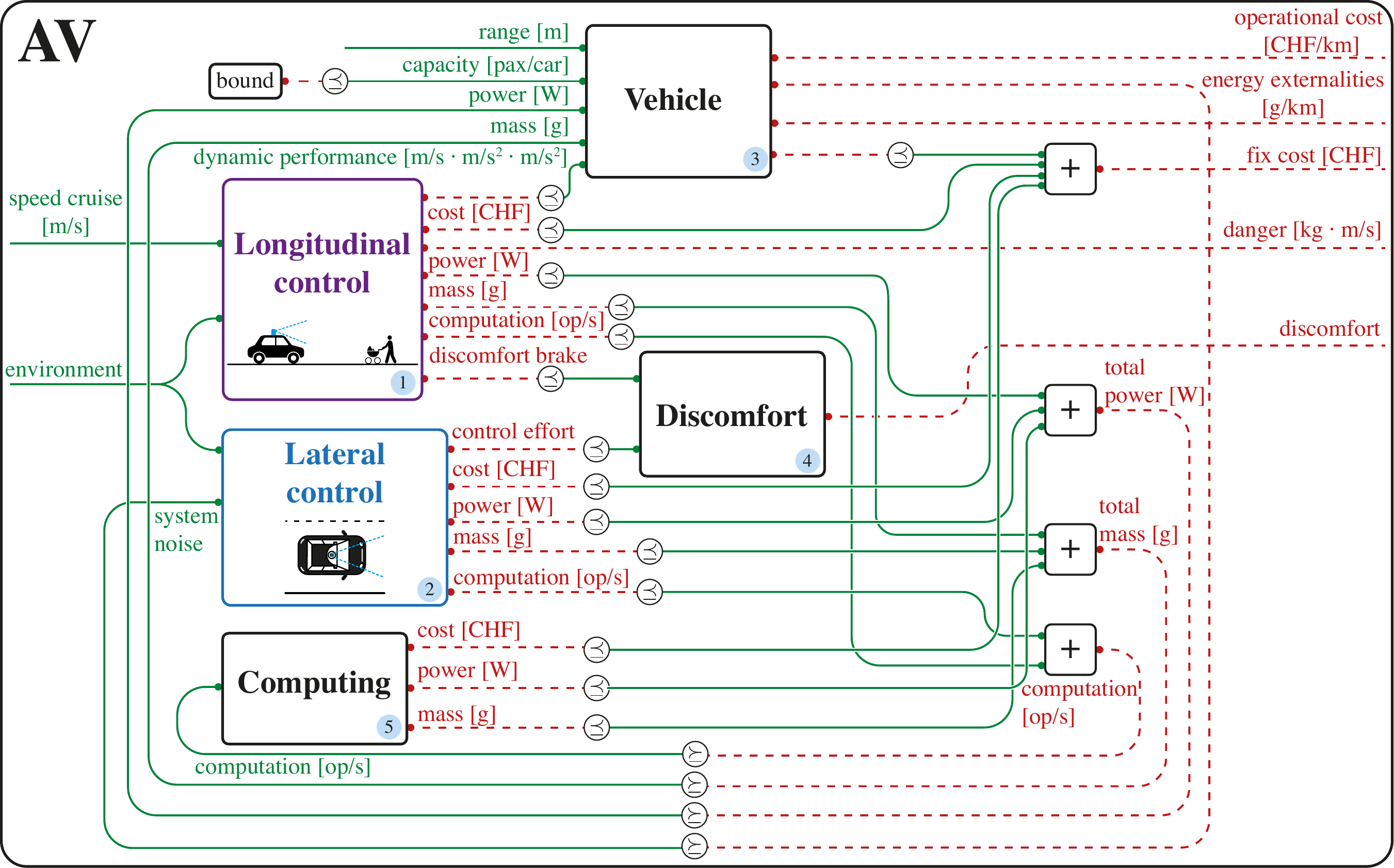}
\caption{Co-design diagram for the design of an \gls{abk:av} which needs to drive safely in a given \Ftext{environment} at given \Ftext{cruise speed} and follow a lane, without hitting obstacles. We choose \Rtext{cost}, \Rtext{externalities}, \Rtext{discomfort} and \Rtext{danger} as the resources to minimize.}
\label{fig:dpgeneralembodied}
\end{subfigure}
\caption{In this work we show how to leverage a monotone theory of co-design to compute the set of optimal solutions for the entire hardware and software stack of an \gls{abk:av}. Each component is obtained following a functional decomposition approach. We illustrate how to obtain design problems from analytical relations, numerical simulations, and catalogues.}
\vspace{-0.5cm}
\end{figure}

\paragraph*{Related work}
The research on embodied intelligence co-design mainly pertains to trade-offs in robotics, sensor and actuator selection and control synthesis. Trade-offs in robotics are studied in~\cite{merlet2005,spielberg2017,lahijanian2018,seok2014,okane2008,karaman2012,bravo2020,nilles2018,ramos2018,collin2019phd}. In particular,~\cite{merlet2005} proposes a formulation for the optimal design of serial manipulators and~\cite{spielberg2017} focuses on articulated robots. The authors of~\cite{lahijanian2018} provide insights on resource-performance trade-offs in mobile robotics,~\cite{seok2014} studies energy-efficient design techniques for legged robots and~\cite{okane2008} assesses the relation between robot sensing and actuation for worst-case scenarios. Furthermore,~\cite{karaman2012,bravo2020} analyze performance limits for robotics tasks, relating them to the complexity of the environment. General robot design schemes are suggested in~\cite{nilles2018,ramos2018,collin2019phd}. Specifically, while~\cite{nilles2018} identifies the principal challenges in modern robotics in \emph{formalization}, \emph{minimality}, \emph{automation} and \emph{integration},~\cite{ramos2018,collin2019phd} present two different frameworks for modular robot design. The selection of sensors and actuators is investigated in~\cite{gupta2006,joshi2008,zhang2020, Shell21, collin2019,guo2019,golovin2010}. Albeit sensor selection problems typically do not admit close form solutions,~\cite{gupta2006,joshi2008} show that for specific instances of the problem there is enough structure for efficient optimization. The authors of~\cite{zhang2020, Shell21} optimally select actuators and sensors to solve arbitrary path planning problems,~\cite{collin2019} focuses on perception architectures for \gls{abk:av} navigation and~\cite{guo2019,golovin2010} focus on large-scale sensor networks. Finally, robotic co-design techniques for control synthesis are examined in~\cite{tanaka2015,tatikonda2004,rosolia2020unified, tzoumas2020,zardiniLCSS2020}. Whilst in~\cite{tanaka2015} the researchers jointly optimize sensor selection and control synthesis, in~\cite{tatikonda2004} an approach to optimal control with communication constraints is explored, and~\cite{rosolia2020unified} propose a hierarchical multi-rate control architecture for actuation and planning. Furthermore, the problem of sensing-constrained LQG control is specifically formalized in~\cite{tzoumas2020} and generally included in robotic platforms co-design problems in~\cite{zardiniLCSS2020}. In summary, current design techniques for embodied intelligence typically have a fixed, problem-specific structure, lacking modular and compositional properties. To the best of our knowledge, no existing framework is able to formalize and solve heterogeneous robot co-design problems, accommodating different modeling techniques.
\paragraph*{Statement of contributions}
In this work we present a methodology to formulate and solve embodied intelligence co-design problems. First, we show how to integrate heterogeneous domains in robotics such as control, actuation, perception and computation. We illustrate how different engineering modeling techniques (analytical, simulation- and catalogue-based) can be incorporated in the same co-design problem. Second, we show how to organize co-design models following the functional decomposition of an embodied intelligence task in sub-tasks, allowing one to capture systemic relations at various abstraction levels. Finally, to showcase our approach, we analyze the example of an \gls{abk:av} driving on a lane and describe how our framework can answer several design queries ranging from sensor selection to control synthesis.
\paragraph*{Organization of the paper}
The remainder of this work is structured as follows. \cref{sec:preliminaries} reviews the mathematical background on which our co-design framework is based. \cref{sec:part_i} presents the functional decomposition approach we use to model robotic systems and illustrates the working principles of our formalism, tailored to the co-design problem of an \gls{abk:av}. \cref{sec:case_study} highlights a sample of the questions we can answer through the proposed methodology and \cref{sec:conclusion} concludes the paper with a discussion and an outlook on future research interests.
\section{Monotone Co-Design Theory}
\label{sec:preliminaries}
We will use basic facts about order theory. A possible reference is Davey and Priestley~\cite{davey2002}.
This section summarizes the key concepts related to the monotone theory of co-design~\cite{censi2015,censi2016}. First, we introduce partially ordered sets, which will be instrumental when defining quantities of interest in design problems.

\begin{definition}[Poset]
A \emph{\gls{abk:poset}} is a tuple~$\tup{P,\preceq_P}$, where~$P$ is a set and~$\preceq_P$ is a partial order, defined as a reflexive, transitive, and antisymmetric relation.
\end{definition}

\begin{definition}[Opposite of a poset]
The \emph{opposite} of a poset~$\tup{P,\preceq_P}$ is the poset $\tup{P^\mathrm{op},\preceq_P^\mathrm{op}}$, which has the same elements as~$P$, and the reverse ordering.
\end{definition}
\begin{definition}[Product poset]
Let~$\tup{P,\preceq_{P}}$ and~$\tup{Q,\preceq_{Q}}$ be posets. Then,~$\tup{P\times Q,\preceq_{P\times Q}}$, with
\begin{equation*}
    \tup{p_1,q_1}\preceq_{P\times Q}\tup{p_2,q_2} \Leftrightarrow p_1\preceq_{P}p_2 \text{ and } q_1\preceq_Q q_2,
\end{equation*}
is the \emph{product poset} of~$\tup{P,\preceq_{P}}$ and~$\tup{Q,\preceq_{Q}}$.
\end{definition}

Given partial orders, one can define monotone maps.

\begin{definition}[Monotone map]
A map~$f\colon P\rightarrow Q$ between two posets~$\langle P, \preceq_P \rangle$,~$\langle Q, \preceq_Q \rangle$ is \emph{monotone} iff~$x\preceq_P y$ implies~$f(x) \preceq_Q f(y)$. Monotonicity is compositional.
\end{definition}

We are now ready to define \glspl{abk:dpi}. 

\begin{definition}[\gls{abk:dpi}]
Consider the \glspl{abk:poset} $\setOfFunctionalities{},\setOfResources{}$, representing \Ftext{functionalities} and \Rtext{resources}, respectively. An \emph{\gls{abk:dpi}} $d$ is a tuple $\tup{\setOfImplementations{d},\prov,\req}$, where $\setOfImplementations{d}$ is the \emph{set of implementations}, and $\prov$, $\req$ are functions mapping $\setOfImplementations{d}$ to $\setOfFunctionalities{}$ and $\setOfResources{}$, respectively:
\begin{equation*}
        \setOfFunctionalities{} \xleftarrow{\prov} \setOfImplementations{d} \xrightarrow{\req} \setOfResources{}.
\end{equation*}
To each \gls{abk:dpi} we \emph{associate} a monotone map $\bar{d}$ given by
\begin{equation*}
    \begin{aligned}
        \bar{d}\colon \setOfFunctionalitiesOp{} \times \setOfResources{} &\to \power{\setOfImplementations{d}}\\
        \langle \F{f}^*,\R{r}\rangle &\mapsto \{i \in \setOfImplementations{d}\colon (\prov(i) \succeq_{\setOfFunctionalities{}}\F{f}) \wedge (\req(i)\preceq_{\setOfResources{}}\R{r})\},
    \end{aligned}
\end{equation*}
where $(\cdot)\op$ reverses the order of a \gls{abk:poset}, and~$\mathcal{P}(\cdot)$ is the powerset. An \gls{abk:dpi} is represented in diagrammatic form as in~\cref{fig:mathcodesign}. The expression $d(\F{f}^*,\R{r})$ gives the set of implementations $S\subseteq \mathcal{I}_d$ for which \Ftext{functionalities} $\F{f}$ are feasible with \Rtext{resources} $\R{r}$. If $\F{f}$ is not feasible with $\R{r}$, $S=\varnothing$.
\end{definition}

\begin{example}[Monotonicity]
\label{ex:computing}
We consider the \gls{abk:dpi} of a \textbf{computing unit} (\cref{fig:dpgeneralembodied}, \circled{5}) by means of the \Ftext{computation} provided and the required \Rtext{cost, mass} and \Rtext{power}. If a set $S=d(\F{f}^*,\R{r})$ of computers require $\R{r}=\tup{\R{\text{mass}}, \R{\text{cost}}, \R{\text{power}}}$ to provide $\F{f}=\F{\text{computation}}$, then they can also provide less computation (reduced performance) $\F{f'}\preceq_\setOfFunctionalities{} \F{f}$, i.e. $S'\supseteq S$. Conversely, if one chooses larger resources $\R{r''}\succeq_\setOfResources{} \R{r}$, the new set of computers will at least still provide $\F{\text{computation}}$, i.e. $S''\supseteq S$.
\end{example}

\begin{definition}[Upper sets]
Consider a poset~$P$. A subset~$S\subseteq P$ is an \emph{upper set} iff~$x\in S$ and~$x\preceq_P y$ implies~$y\in S$.
\end{definition}
\begin{definition}[Antichain]
Consider a poset~$P$. A subset~$S\subseteq P$ is an antichain iff no elements are comparable. In other words, iff for~$s_1,s_2\in S$,~$s_1\preceq_P s_2$ implies~$s_1=s_2$.
\end{definition}
We denote the set of all antichains in~$P$ by~$\mathsf{A}P$. 

\begin{definition}
\label{def:map_h}
Given an \gls{abk:dpi} $d$, we can define the monotone maps $h_d$ and $h'_d$. The map $h_d\colon \setOfFunctionalities{}\to \mathcal{A} \setOfResources{}$ assigns a functionality $\F{f} \in \setOfFunctionalities{}$ to the \emph{minimum antichain} of resources $\mathcal{A}\setOfResources{}$ which provide $\F{f}$. The map $h_{d}'\colon \setOfResources{}\to \mathcal{A}\setOfFunctionalities{}$ maps a resource $\R{r}\in \setOfResources{}$ to the \emph{maximum antichain} of functionalities provided by $\R{r}$. To solve co-design problems, one has to determine these maps, relying on Kleene's fixed point theorem~\cite[Section VIII]{censi2015}. 
\end{definition}

\begin{figure}[t]
\begin{center}
\begin{subfigure}[b]{\linewidth}
    \centering
    \scalebox{1}{\begin{tikzpicture}[DP]
            \node[dp={2}{2}] (cnt) {MDPI};
            \draw[runconn, runame={poset of resources}, relres=above,posres=1.8] (cnt_res1){};
            \draw[runconn, runame={}, relres=above,posres=0.9] (cnt_res2){};
            \draw[funconn, funame={poset of functionalities},relfun=above,posfun=2] (cnt_fun1){};
            \draw[funconn, funame={},relfun=above,posfun=1.15] (cnt_fun2){};
\end{tikzpicture}}
    \subcaption{A design problem is a monotone relation between \Ftext{functionalities} and \Rtext{resources}.}
    \label{fig:mathcodesign}
\end{subfigure}
\begin{subfigure}[b]{0.3\columnwidth}
\centering
\scalebox{0.8}{\begin{tikzpicture}[DP]
    \node[dp={1}{1}] (f) {$f$};
    \node[dp={1}{1}, right=1cm of f] (g) {$g$};
    \draw[rconn, rcname={}, fcname={}] (f_res1)  to (g_fun1);
    \draw[runconn, runame={}] (g_res1);
    \draw[funconn, funame={}] (f_fun1);
\end{tikzpicture}}
\subcaption{Series composition.}
\end{subfigure}
\begin{subfigure}[b]{0.3\columnwidth}
\centering
\scalebox{0.8}{\begin{tikzpicture}[DP]
    \node[dp={1}{1}] (f) {$f$};
    \node[dp={1}{1}, below=0.3cm of f] (g) {$g$};
    \draw[runconn, runame={}] (f_res1){};
    \draw[runconn, runame={}] (g_res1){};
    \draw[funconn, funame={}] (f_fun1){};
    \draw[funconn, funame={}] (g_fun1){};
\end{tikzpicture}}
\subcaption{Parallel composition.}
\end{subfigure}
\begin{subfigure}[b]{0.3\columnwidth}
\centering
\scalebox{0.8}{\begin{tikzpicture}[DP]
    \node[dp={2}{2}] (f) {$f$};
    \draw[runconn, runame={}] (f_res1){};
    \draw[funconn, funame={}] (f_fun1){};
    \draw[rconn,rcname={},fcname={},feedback=1,loos=1.5] (f_res2) -| ($(f)-(0,4)$) |- (f_fun2);
\end{tikzpicture}}
\subcaption{Loop composition.}
\end{subfigure}
\label{fig:diagrams}
\caption{(a) A single \gls{abk:dpi} is characterized by \Ftext{functionalities} and \Rtext{resources} and (b-d) multiple \glspl{abk:dpi} can be composed in different ways.}
\label{fig:dpcompositions}
\end{center}
\vspace{-0.5cm}
\end{figure}
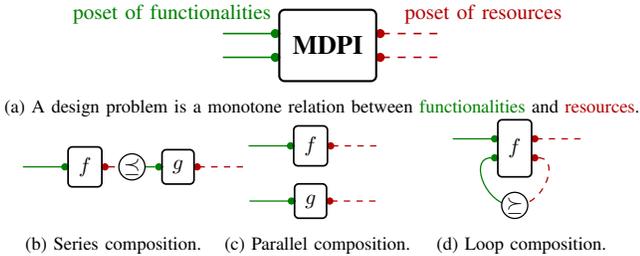

Individual \glspl{abk:dpi} can be composed in several ways to form a co-design problem (\cref{fig:dpcompositions}), and design problems are closed under composition. Series composition describes the scenario in which the functionality of an \gls{abk:dpi} becomes the resource of another \gls{abk:dpi}. For instance, the energy produced by a battery is required by actuators. Parallel composition corresponds to processes happening together. Finally, loop composition describes feedback\footnote{It can be proved that the formalization of feedback turns the category of \glspl{abk:dpi} into a traced monoidal category~\cite{fong2019}. 
}. As shown in~\cite{censi2015,fong2019}, the presented composition operations preserve monotonicity and thus all related algorithmic properties.

\section{A functional decomposition approach}
\label{sec:part_i}
\subsection{Embodied intelligence co-design patterns}
To effectively co-design a robotic platform, it is appropriate to create a \emph{functional decomposition} of the platform's task. %
The guiding example in this paper is that of an \gls{abk:av}, but we argue that the same principle can be generalized to embodied intelligence. We consider as the main function \emph{urban driving}~(\cref{fig:decomposition}). We zoom in on the \emph{follow lane} function (parallel functions would be \emph{handle intersections}, \emph{pick up passenger}, etc).
One can decompose lane-following into two more specific functions: \emph{lateral control} (i.e., stay in a lane) and \emph{longitudinal control} (i.e., accelerate and brake in presence of obstacles). For these two functions we provide
a reasonably detailed analysis. Note that there is no limitation for our method to decompose a system into an arbitrary number of functions.

In the following we refer to ``functions'' (in the systems engineering nomenclature) as ``tasks'', since the term ``functions'' clashes with co-design ``functionalities''.

\paragraph*{Task abstraction} The output of functional decomposition is a collection of sub-tasks, which in co-design can commonly be abstracted as \glspl{abk:dpi}~(\cref{fig:dptask}). In particular, embodied intelligence tasks share a common structure, constituted of task-specific functionalities, such as \Ftext{performance} and system-level functionalities shared with other components such as \Ftext{scenarios/environments} (e.g., time of the day, density of obstacles). Furthermore, we identified \Rtext{cost} (in \unit[]{CHF}), \Rtext{power} (in \unit[]{W}), \Rtext{computation effort} (in \unit[]{op/s}) and \Rtext{mass} (in \unit[]{g}) as the four main resources which ``need to be paid'' to achieve a particular desired task performance in any embodied intelligence scenario. 
These four resources are the result of a simplification and models can be refined at will, e.g., by distinguishing operation, maintenance and fixed costs, or by analyzing CPU and GPU computation separately.
\paragraph*{From function decomposition to co-design diagrams}
Given specific task abstractions, one can convert a functional decomposition diagram into a co-design diagram (\cref{fig:dpfundeco}). To do so, the specific \Ftext{scenarios/environments} are fed into each sub-task and a general \Ftext{task performance} is assigned to the problem. In particular, the functional decomposition block has knowledge of the task decomposition logic and knows for each task performance level which combinations of performance levels are required. The \Rtext{resources} required by the different sub-tasks are generated independently and then ``summed''\footnote{Or, in general, an associative operation is applied, such as $\max{}$ or +.}. We note the compositionality property of this formalization: the resulting diagram has the same interface as that of a task, meaning that a composite task is a task. 

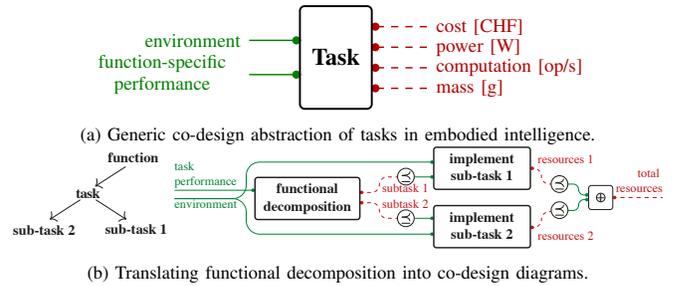
\begin{figure}[t]
\begin{subfigure}[b]{\linewidth}
    \centering
    \scalebox{0.9}{\begin{tikzpicture}[DP, dp port sep = 1.6]
            \node[dp={2}{4}] (cnt) {Task};
            \draw[funconn, funame={environment}, relfun=left] (cnt_fun1){};
            \draw[funconn, funame={\begin{tabular}{c}function-specific\\ performance\end{tabular}}, relfun=left,posres=0] (cnt_fun2){};
            \draw[runconn, runame={cost \unit[]{[CHF]}},relres=right] (cnt_res1){};
            \draw[runconn, runame={power \unit[]{[W]}},relres=right] (cnt_res2){};
            \draw[runconn, runame={computation \unit[]{[op/s]}},relres=right] (cnt_res3){};
            \draw[runconn, runame={mass \unit[]{[g]}},relres=right] (cnt_res4){};
\end{tikzpicture}}
    \subcaption{Generic co-design abstraction of tasks in embodied intelligence.}
    \label{fig:dptask}
\end{subfigure}
\begin{subfigure}[b]{\columnwidth}
\centering
\includegraphics[width=\linewidth]{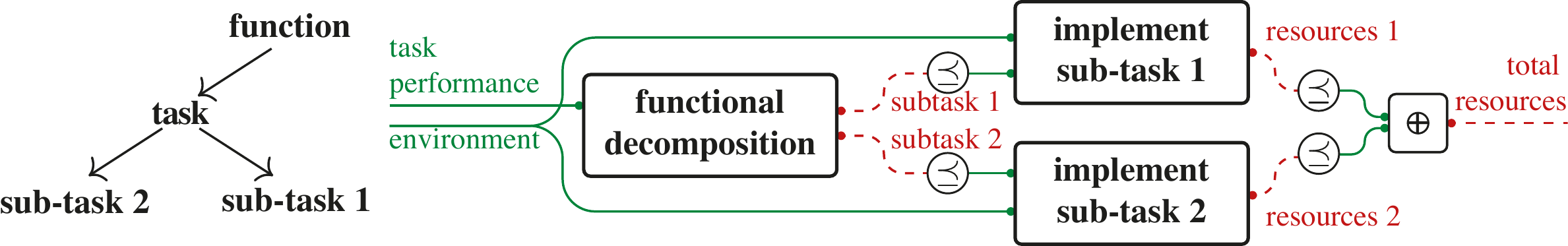}
\subcaption{Translating functional decomposition into co-design diagrams.}
\label{fig:dpfundeco}
\end{subfigure}
\caption{Functional decomposition provides us with sub-tasks, each of which we can model as an \gls{abk:dpi} with \Ftext{environment} and \Ftext{performance} as functionalities and \Rtext{cost}, \Rtext{power}, \Rtext{computation power} and \Rtext{mass} as resources. The interconnection of several tasks composing a functional decomposition is an \gls{abk:dpi} and has the interface of a task, highlighting the compositionality properties of our approach. Resources can be merged together via an associative operation.}
\label{fig:dpembodiedint}
\vspace{-0.5cm}
\end{figure}

\paragraph*{From data flow to co-design diagrams}
To understand the co-design approach, one needs to distinguish logical dependencies and data-flow (\cref{fig:logicvsdata}). By considering logical dependencies (\cref{fig:logicaldeps}), we realize that the purpose of a robotic task implementation is to provide \emph{decision making}, which, to be produced, needs \emph{state information}. The latter is estimated through \emph{sensing data}, provided by a \emph{sensor}. The actual data flow in robotic tasks develops in the reverse direction (\cref{fig:dataflowdeps}). Co-design diagrams are a formalization of logical dependencies and are not to be understood as signal-flow diagrams. In the following we present the formalization of the co-design problem of an \gls{abk:av}, highlighting the properties of the monotone co-design theory.

\begin{figure}[t]
\begin{subfigure}[b]{\columnwidth}
\centering
\begin{tikzcd}[font=\scriptsize]
    \begin{tabular}{c} decision \\ making \end{tabular}\arrow{r}{\text{requires}} &
    \begin{tabular}{c} state \\ estimation \end{tabular}\arrow{r}{\text{requires}}&
    \begin{tabular}{c} sensing \\ data \end{tabular}\arrow{r}{\text{requires}}&
    \begin{tabular}{c} sensor \end{tabular}
\end{tikzcd}
\subcaption{Logical dependency between decision making, state estimation and sensing data.}
\label{fig:logicaldeps}
\end{subfigure}\\[+5pt]
\begin{subfigure}[b]{\columnwidth}
\centering
    \begin{tikzpicture}
    \node[block, minimum width=4.75cm, minimum height= 2cm] (com) at (0,0) {\begin{tabular}{c}computer \\ \vspace{1.5cm} \end{tabular}};
    \node[block,font=\scriptsize, right=1cm of com.east](sen) {sensor};
    \node[block,draw=none,fill=blue!30,align=center,font=\scriptsize, left=5pt of com.east] (est) {\begin{tabular}{c}estimation \\ node \end{tabular}};
    \node[block,fill=blue!15,draw=none,align=center,font=\scriptsize, below=-1pt of est.south] (est_alg) {\begin{tabular}{c}estimation \\ algorithm \end{tabular}};
    \node[block,draw=none,fill=blue!30,align=center,font=\scriptsize,right=5pt of com.west] (cont) {\begin{tabular}{c}controller \\ node \end{tabular}};
    \node[block,fill=blue!15,draw=none,align=center,font=\scriptsize, below=-1pt of cont.south] (cont_alg) {\begin{tabular}{c}controller \\ algorithm \end{tabular}};
    \tikzset{>=latex}
    \draw[->,thick] (est.west) -- (cont.east) node[above, pos=0.5,font=\scriptsize]{\begin{tabular}{c}state\\ estimate \end{tabular}};
    \draw[->,thick] (cont.west) --++ (-1.1,0) node[above, pos=0.7,font=\scriptsize]{command};
    \draw[->,thick] (sen.west) -- (est.east) node[above, pos=0.4,font=\scriptsize]{\begin{tabular}{c}sensing\\ data \end{tabular}};
    \end{tikzpicture}
\subcaption{Data-flow between sensing data, state estimation and decision making.}
\label{fig:dataflowdeps}
\end{subfigure}
\caption{To model embodied intelligence tasks, we first consider (a) logical dependencies. The purpose of the implementation of a task is to provide decision making, which relies on state information. To estimate the state, one needs sensing data, produced by a sensor. (b) The data-flow develops in the opposite direction.}
\label{fig:logicvsdata}
\end{figure}
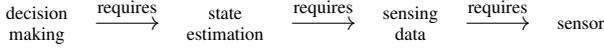
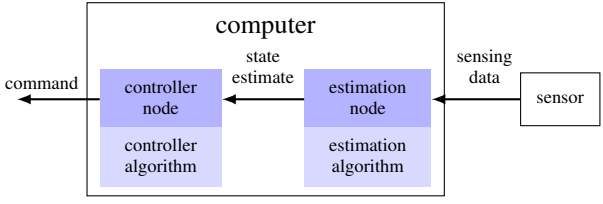
\subsection{Modeling lateral control}
\label{sec:part_ii}
We consider the task of lane-following as the combination of the lateral and longitudinal control tasks. In this paragraph we focus on the co-design of lateral control (\cref{fig:dplateral}).
We explain each block of the diagram in turn, by describing its working principles and how to obtain its model.

\begin{figure}[t]
    \centering
    \includegraphics[width=\linewidth]{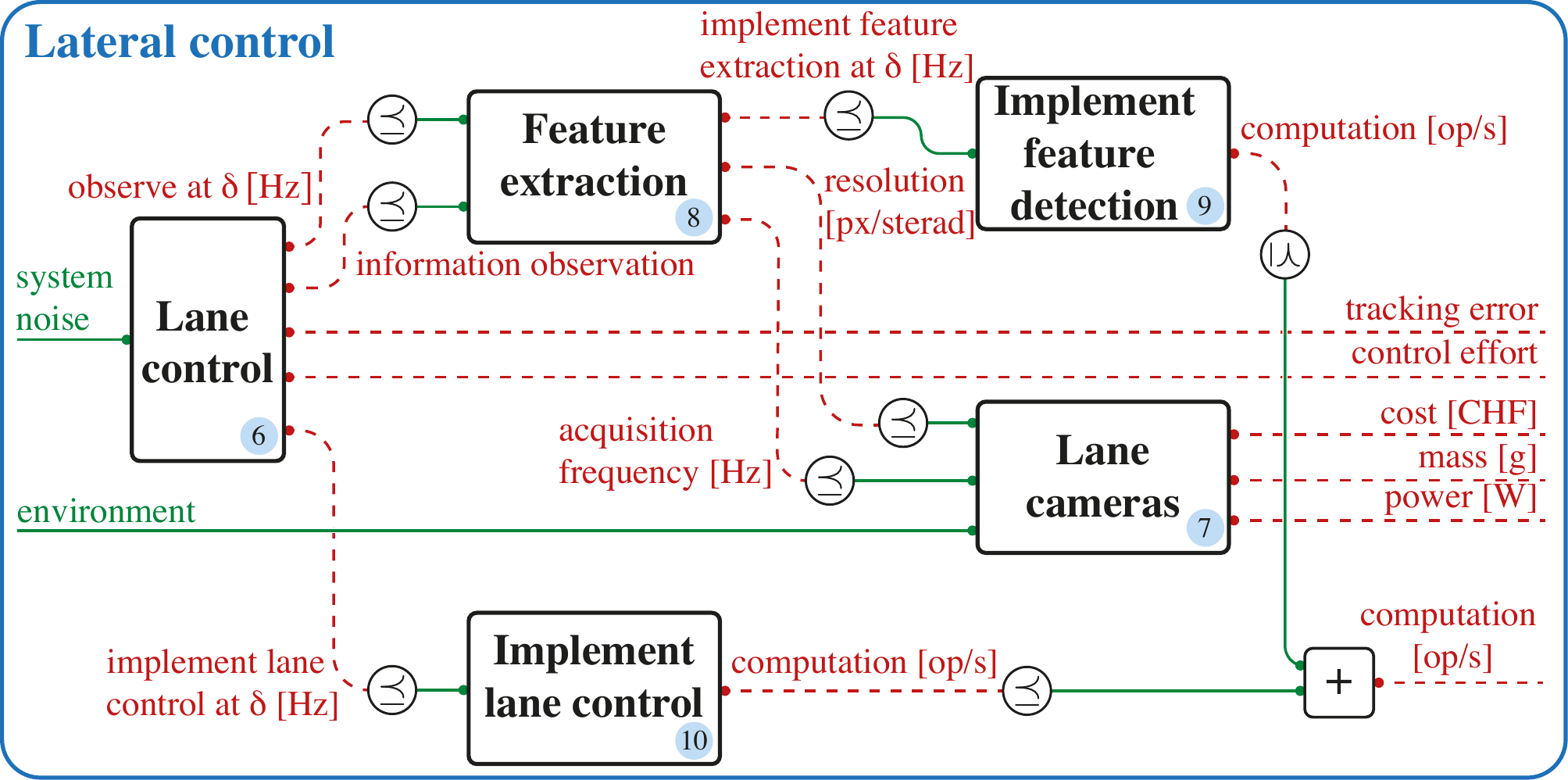}
    \caption{Co-design highlights logical dependencies. In this work we consider lateral control as an \gls{abk:dpi}, involving the design of optimal control strategies and feature detection algorithms, together with sensor selection. Resources are \Rtext{cost}, \Rtext{mass}, \Rtext{power}, \Rtext{computation}, \Rtext{control effort} and \Rtext{tracking error}.}
    \label{fig:dplateral}
\end{figure}

\emph{Block \textbf{Lane Control}}:
Given the task of the vehicle to align itself with the lane, we define the \gls{abk:av} configuration as~$\bm{x}_t=\tup{\theta_t,y_t}$, where~$\theta_t$ is the heading of the \gls{abk:av} and~$y_t$ is its relative lateral position with respect to the center of the lane. The desired lane-aligned configuration at time~$t$ is denoted by~$\bm{x}_t^\mathrm{g}=\tup{\theta_t^\mathrm{g},y_t^\mathrm{g}}$ and the control input is the steering  torque~$\tau_t$.  
We assume that from the sensor we can have Gaussian observations of the state $\bm{o}_t=\bm{x}_t-\bm{x}_t^{\mathrm{g}}+\bm{v}_t$, where~$\bm{v}_t$ is white Gaussian noise with covariance~$\mat{\Sigma}_{\bm{v}}$.
As we show in~\cite{zardiniLCSS2020}, this problem can be solved with LQG control, choosing to minimize the objective~$\textstyle{\lim_{T\to \infty}} \frac{1}{T}\mathbb{E}\{\smallint_{0}^{T} \left(\left( \bm{x}_t^\intercal \alpha \mat{Q}_0 \bm{x}_t\right)+\left( r_0 \tau_t^2/\alpha\right)\right) \D t\}$. This can be formalized as the \textbf{lane control} \gls{abk:dpi} \circled{6} (\cref{fig:dplateral}), in which the functionality is the ability of the control to handle a given \Ftext{system noise} and the resources are the needs to obtain the \Rtext{observations} at a certain frequency with given \Rtext{precision} and to \Rtext{implement the optimal control law} at a specific \Rtext{frequency} (in \unit[]{Hz}), providing \Rtext{control effort}~$J_\mathrm{eff}=\textstyle{\lim_{t\to \infty}}\mathbb{E}\{r_0\tau^2_t\}$ and \Rtext{tracking error}~$J_\mathrm{track}=\textstyle{\lim_{t\to \infty}}\mathbb{E}\{\bm{x}_t^\intercal \mat{Q}_0\bm{x}_t\}$. The implementations the designer can choose are the different cost weights, parametrized by~$\alpha$. 
\emph{Obtaining the model}: We show in~\cite{zardiniLCSS2020} how the nature of the problem allows to obtain the optimal solutions for the \gls{abk:dpi} by solving specific Riccati equations.

\emph{Block \textbf{Lane Cameras}}:
We need to define a relation between the accuracy of the sensing and the physical sensors. Measurements are provided at a given \Ftext{frequency} and with a specific \Ftext{resolution} (in \unit[]{px/sterad}) by \textbf{lane cameras} \circled{7}, which have a \Rtext{cost}, \Rtext{mass} and \Rtext{power consumption}.
\emph{Obtaining the model}: This is obtained straight from camera catalogues.

\emph{Block \textbf{Feature Detection}}:
A feature detection algorithm \circled{8} processes the measurements providing the lane control \gls{abk:dpi} with observations at a certain \Ftext{frequency} and with a  certain \Ftext{precision}. 
\emph{Obtaining the model}: This is the realm of photogrammetry. We need to answer questions such as ``what \Rtext{resolution} (in \unit[]{px/sterad}) is needed to achieve a certain feature detection \Ftext{accuracy}?''.

\emph{Blocks \textbf{Algorithms Implementation}}:
Finally, it is necessary to choose the implementation of the actual feature detection and lane control algorithms (\circled{9} and \circled{10}). For each of these we have an \gls{abk:dpi} characterized by a catalogue of algorithms, each requiring different \Rtext{computation power}. 
\emph{Obtaining the model}: This is a benchmarking exercise.
An excellent example of how to create benchmarking catalogues for algorithms, going as deep as to also 
search over the compiler flags, is given by \emph{SLAMBench}~\cite{nardi15} and the successive papers by Nardi and collaborators. 
For perception problems which cannot be adequately modeled by analytical photogrammetry relations,
it also makes sense to not only vary implementation details (e.g., compiler flags) but also 
algorithm parameters. In that case, benchmarking would include the scope of the last two blocks together.

\emph{Entire \textbf{lateral control} diagram}:
The general \textbf{lateral control} co-design diagram \circled{2} (\cref{fig:dpgeneralembodied}) includes robustness to specific \Ftext{system noise} and \Ftext{environment} and requires the common resources \Rtext{cost}, \Rtext{mass}, \Rtext{power} and \Rtext{computation power}, together with \Rtext{tracking error} and \Rtext{control effort}, which will be important when implementing safety and control metrics. 
If the reader agrees that each block is an \gls{abk:dpi} then their composition is an \gls{abk:dpi}. One should already notice that the \textbf{lateral control} design problem involves the co-design of hardware and software components. 

\subsection{Modeling longitudinal control}
As highlighted above, lateral control can be modeled as a co-design problem using analytical solutions of optimal control problems. For the longitudinal control sub-task, however, we want to showcase the ability of our framework to handle cases in which co-design relations are not directly available in analytical form, to the point at which one has to rely on tools such as numerical simulation (\cref{fig:dplongcontrol}). The \gls{abk:av} is required to brake in time in the presence of obstacles and to guarantee a desired \Ftext{cruise speed}. The \gls{abk:av} is characterized by a dynamic performance~$\tup{v_\mathrm{max},a_\mathrm{max},a_\mathrm{min}}$, where~$v_\mathrm{max}$ is the maximum vehicle's achievable speed and~$a_\mathrm{max}$, $a_\mathrm{min}$ are its maximum acceleration and deceleration. These parameters depend on the chosen vehicle type (e.g., on the propulsion system). The vehicle's longitudinal dynamics are~$\D x_t=v_t \D t$,~$\D v_t = a_t \D t$, where~$a_t$ and~$v_t$ represent the vehicle's acceleration and velocity. 

\emph{Block \textbf{Longitudinal sensing}}: The \gls{abk:av} is equipped with sensors, which provide obstacle detections along the road. It has already been observed that sensors can be ordered by their ability to discriminate states~\cite{lavalle2012}. In our work, each sensor is characterized by its \Ftext{sensing performance}, expressed as a tuple~$\tup{\F{\fp(d)},\F{\fn(d)},\F{\acc(d)}}$, where~$\fp,\fn \colon \mathbb{R}_{\geq 0} \to \mathbb{R}_{[0,1]}$ represent false positives (i.e., given an environment without obstacles, the probability of detecting one) and false negatives (i.e., assuming the presence of an obstacle, the probability of not detecting it) curves as a function of distance from the obstacle and with~$\acc \colon \mathbb{R}_{\geq 0} \to \mathbb{R}_{\geq 0}$ denoting the sensing accuracy (range) as a function of distance from the obstacle.
These curves can be obtained is different ways, as explained in \cref{sec:case_study}. Note that to consider the curves as functionalities in the \textbf{longitudinal sensing} \gls{abk:dpi} \circled{12}, we compare them in \glspl{abk:poset} using the point-wise order, and combine the \glspl{abk:poset} by taking their product, resulting in the sensor performance \gls{abk:poset}. In~\cref{fig:sensorslong}, one can see examples of these curves together with the corresponding Hasse diagrams for the \glspl{abk:poset}. The Hasse diagram can be understood as follows. In the false positives \gls{abk:poset}, sensor Pointgrey is dominated by Ace5gm, Ace251gm and, by transitivity, by Ace13gm. The detections also depend on the \Ftext{environment} in which the task needs to be solved. This includes the time of the day as well as the density of obstacles on the road. Furthermore, a sensor provides measurements at a certain \Ftext{acquisition frequency} and has specific \Rtext{latency} (in \unit[]{s}), \Rtext{cost}, \Rtext{mass} and \Rtext{power}.
\emph{Obtaining the model}: One obtains sensor specifications from catalogues and detection properties from benchmarking. 

\begin{figure}[t]
\begin{center}
\includegraphics[width=\columnwidth]{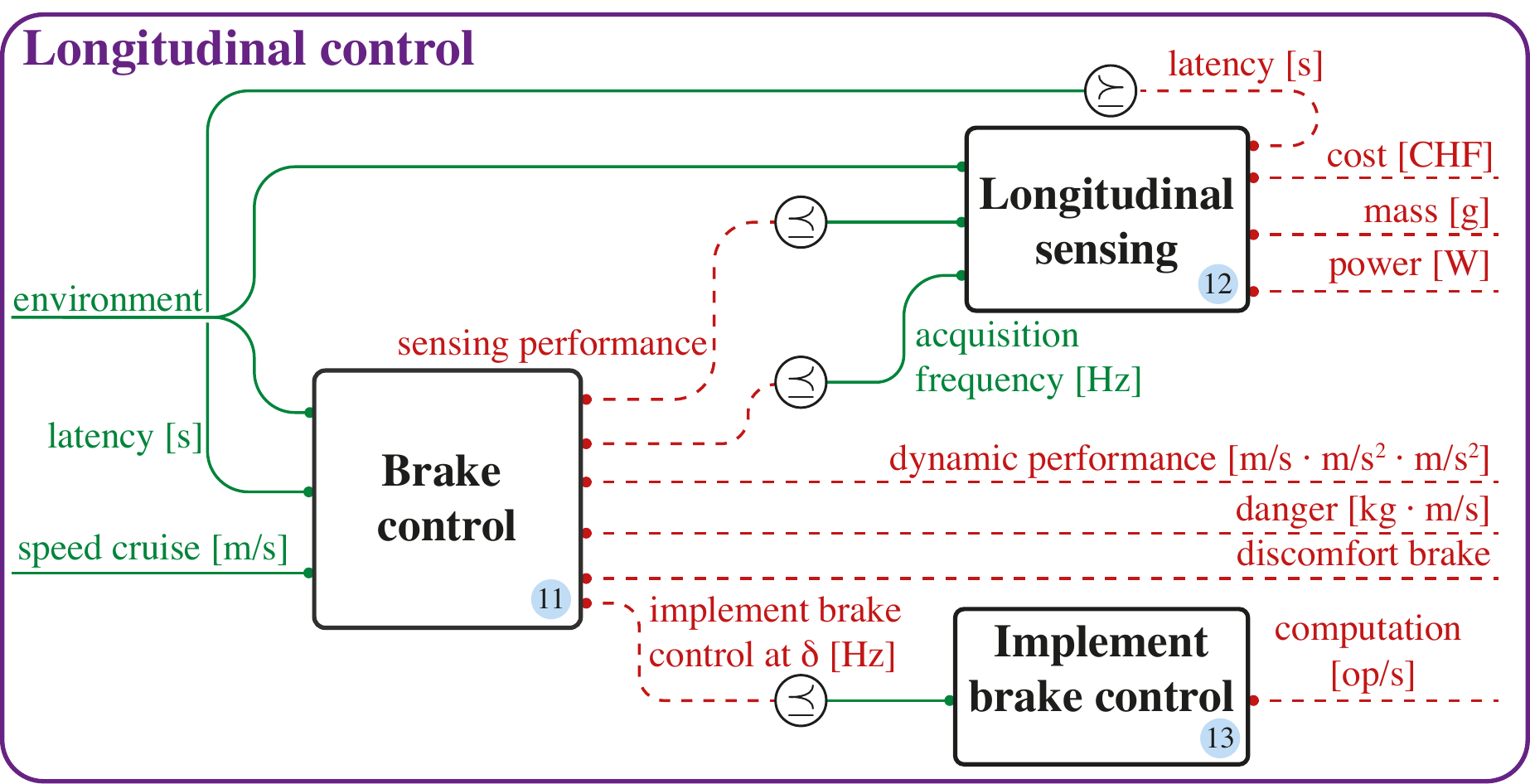}
\caption{The longitudinal control \gls{abk:dpi} consists of a brake control, a longitudinal sensing and an implementation block. It provides the \gls{abk:av} with the ability of reaching a \Ftext{cruise speed} in a given environment, requiring \Rtext{cost}, \Rtext{mass}, \Rtext{power}, \Rtext{dynamic performance}, \Rtext{danger}, \Rtext{discomfort} and \Rtext{computation}.}
\label{fig:dplongcontrol}
\end{center}
\vspace{-0.5cm}
\end{figure}

\begin{figure}[t]
\begin{center}
\includegraphics[width=\columnwidth]{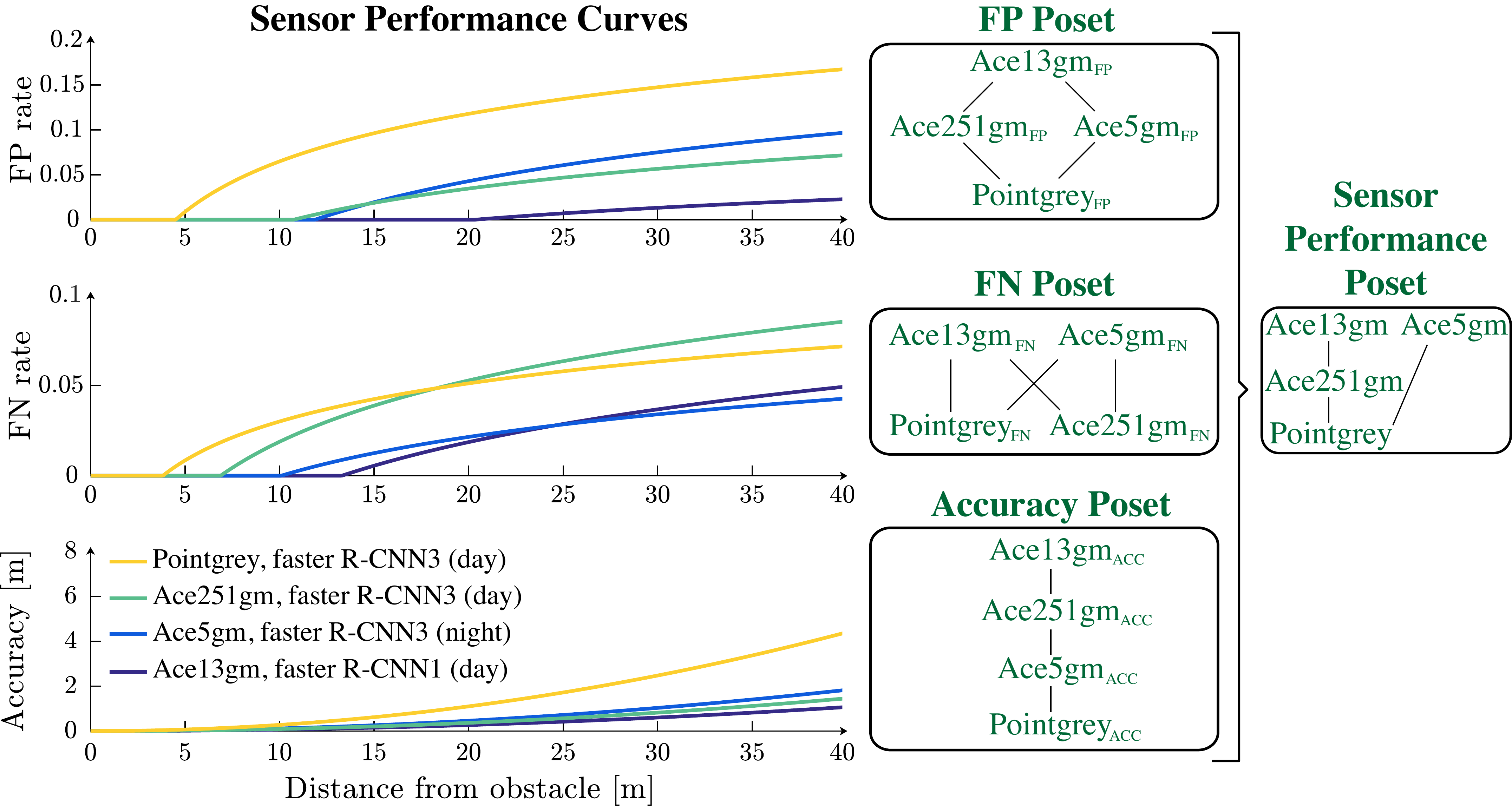}
\caption{We consider a catalogue of sensors, characterized by \Rtext{cost}, \Rtext{mass}, \Rtext{power consumption}, \Rtext{latency}, \Ftext{acquisition frequency} and their \Ftext{performance}, represented through the three curves of false positives, false negatives and accuracy. The curves are ordered in a \gls{abk:poset} and the sensor performance product \gls{abk:poset} is computed.}
\label{fig:sensorslong}
\end{center}
\vspace{-0.5cm}
\end{figure}

\emph{Block \textbf{Brake control}}: Based on the generative model of the measurements, we produce a Bayesian estimate of the probability of having a pedestrian at each distance $d$ and denote it by $f(d)$.
Because it is not possible to derive a closed form solution for this POMDP, we choose the parametrized control law in~\cref{alg:long_cont}.
\begin{footnotesize}
\begin{algorithm}
\caption{Longitudinal controller} \label{alg:long_cont}
\begin{algorithmic}[1]
    \State \textbf{Input:} Current speed $v_t$, Bayesian estimate $f(x)$.
    \State \textbf{Output:} Control input $a_t \in \{a_\mathrm{min}, a_\mathrm{max}, 0\}$.
    \State \textbf{Parameters:} Speed $v_\mathrm{cruise}$, braking distance $d_{\mathrm{crit},t}$, threshold $\Theta$, dynamic performance~$\tup{v_\mathrm{max},a_\mathrm{max},a_\mathrm{min}}$.
            \Procedure{}{}
                \State $d_{\mathrm{crit},t}$ = $ v_t^2/(2 \cdot a_\mathrm{min})$ \;
                \If {$\textstyle{\left(\int_{0}^{d_{\mathrm{crit},t}} f(x) \D x\right)}$ $\geq$ $\Theta$} $a_t = a_\mathrm{min}$ \;
                \ElsIf {$v_t < v_{cs}$} $a_t = a_\mathrm{max}$ \;
                \Else { $a_t = 0$} \;
                \EndIf
            \EndProcedure
        \end{algorithmic}
\end{algorithm}
\end{footnotesize}

Doing so, we can write the \textbf{brake control} problem as an \gls{abk:dpi} \circled{11} (\cref{fig:dplongcontrol}) in which functionalities are the provided \Ftext{cruise speed} (i.e., the performance, in \unit[]{km/h}) and the handled \Ftext{sensing latency} and \Ftext{environment}. The resources are the \Rtext{sensing performance}, the sensing \Rtext{acquisition frequency}, the \Rtext{computation power} needed to execute the control law and the \Rtext{dynamic performance} of the vehicle. Furthermore, we measure the performance of the longitudinal control action by means of \Rtext{discomfort} and \Rtext{danger}, defined as follows. Given a time horizon $T$, \Rtext{discomfort} is expressed as $\textstyle{\mathsf{dis}=\int_0^T \vert a_t\vert \D t}$ and penalizes changes in acceleration~\cite{reschka2012}. \Rtext{Danger} captures both the \emph{probability} and the \emph{impact} of failures and is expressed as the product of the probability of hitting an obstacle and the momentum of the collision (in \unit[]{kg$\cdot$m/s}).
\emph{Obtaining the model}: Given the sensing model and the controller parameters, the \textbf{brake control} \gls{abk:dpi} can be modeled by running numerical simulations. This approach has two fundamental advantages. First, the simulations can be run in parallel, as they do not depend on each other. Second, the general co-design optimization can be run with incomplete simulation results, obtaining reduced levels of accuracy for the co-design solutions. It can actually be shown that the accuracy of the solutions of the co-design problem is monotone with the number of simulations one has~\cite{censi2017}. Leveraging such properties is important material for future research.

\emph{Block \textbf{implementation brake control}}: Brake control needs to be actually \textbf{implemented} \circled{13} (\cref{fig:dplongcontrol}).
\emph{Obtaining the model}: Different implementations of the control algorithms have different properties in terms of control update \Ftext{frequency} and required \Rtext{computational effort}

\emph{Entire \textbf{longitudinal control} diagram}: By interconnecting the aforementioned blocks, one obtains the co-design diagram for the longitudinal control of an \gls{abk:av} (\cref{fig:dplongcontrol}). The general \textbf{longitudinal control} co-design diagram \circled{1} (\cref{fig:dpgeneralembodied}) features the first loop through \Rtext{latency} and is subject to the \Ftext{environment} and desired \Ftext{cruise speed} and requires \Rtext{cost}, \Rtext{mass}, \Rtext{power}, \Rtext{dynamic performance}, \Rtext{danger}, \Rtext{discomfort} and \Rtext{computation}.

\subsection{Composing the full diagram}
In the previous sections we detailed the modeling of the \textbf{lateral control} and \textbf{longitudinal control} \glspl{abk:dpi}. Following the principle of functional decomposition, we can now interconnect these two components with the rest of the system, obtaining the general co-design diagram reported in~\cref{fig:dpgeneralembodied}. 
\emph{Block \textbf{Computing Unit}}: This problem includes the design of the \textbf{computing} unit \circled{5} (\cref{ex:computing}), which needs to provide \Ftext{computation power} required by all the processes we have presented and which requires \Rtext{power} and has a \Rtext{mass} and a \Rtext{cost}. As illustrated in~\cref{fig:dpgeneralembodied}, the feedback of computation power constitutes the second feedback loop of our co-design diagram.
\emph{Obtaining the model}: The \textbf{computing unit} can be modeled through computer catalogues.

\emph{Block \textbf{Vehicle}}:
Furthermore, the general co-design diagram includes the \textbf{vehicle} \gls{abk:dpi} \circled{3} (\cref{fig:dpgeneralembodied}), which represents the mechanical part of the system. This can be formulated as a \gls{abk:dpi} which provides certain \Ftext{dynamic performance} and \Ftext{range} (in \unit[]{m}) and has a certain \Ftext{capacity} (in \unit[]{pax/car}), requiring both operational and fix \Rtext{costs} and \Rtext{energy externalities} (in \unit[]{g/km}). Specifically, the vehicle has the ability to provide \Ftext{power} which allows \textbf{computing}, \textbf{longitudinal control} and \textbf{lateral control} to happen (third feedback loop) and to carry extra \Ftext{mass}, arising from the sum of the mass of the sensors and computing unit (fourth feedback loop). Both increased \Ftext{mass} and \Ftext{power} reduce the vehicle \Ftext{range}. Each vehicle is characterized by a \Ftext{system noise}, which is fed back into the \textbf{lateral control} block (fifth feedback loop). 
\emph{Obtaining the model}: This model can be extracted from catalogues (e.g., by switching propulsion systems and chassis).

\emph{Block \textbf{Discomfort}}:
Finally, a \textbf{discomfort} \gls{abk:dpi} \circled{4} (\cref{fig:dpgeneralembodied}) joins the \Rtext{discomfort} arising from the \textbf{longitudinal control} and from the \textbf{lateral control} (in terms of \Rtext{control effort}) to produce the \Rtext{discomfort} performance metric. 
\emph{Obtaining the model}: One can combine discomfort metrics arising from \textbf{lateral} and \textbf{longitudinal} controls using different assumptions, generating particular discomfort models.

\begin{figure}[t]
    \centering
    \includegraphics[width=0.85\columnwidth]{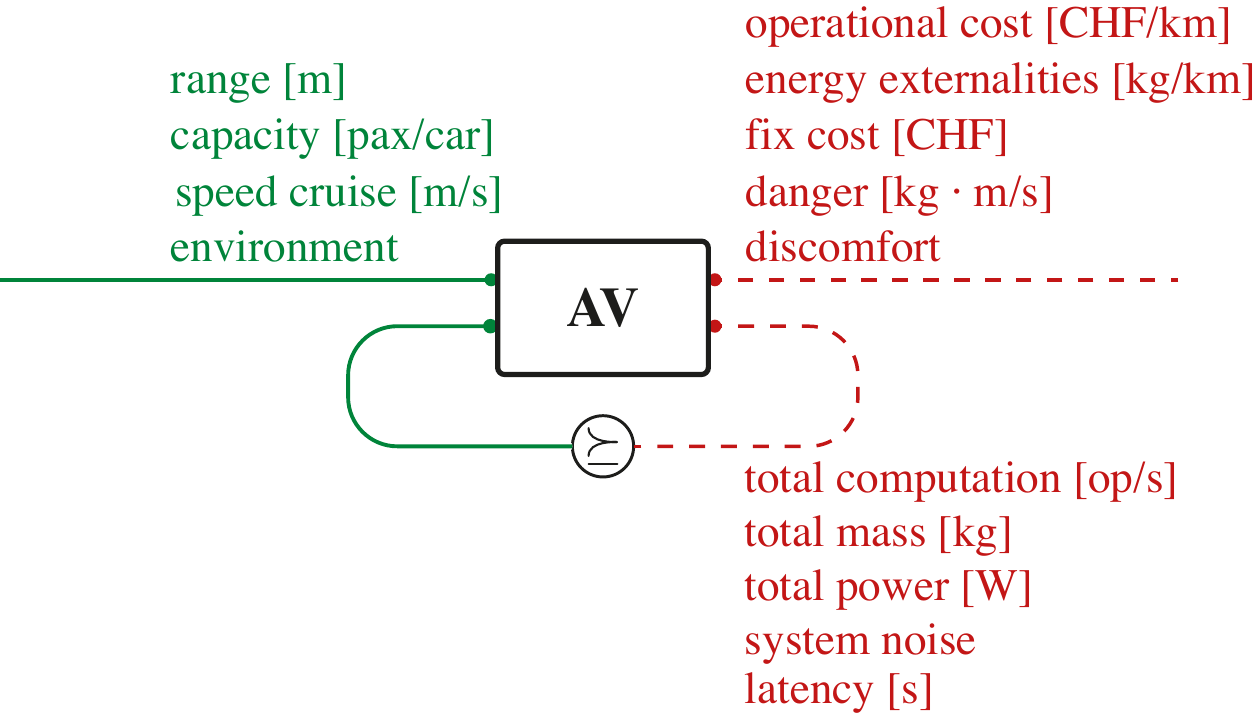}
    \caption{The canonical form distinguishes between \Rtext{resources} which are fed back as \Ftext{functionalities} and those who are not. The number of loops provides an upper bound for the complexity of the design solution~\cite{censi2017}.}
    \label{fig:dpcanonical}
\end{figure}

\emph{General \textbf{\gls{abk:av}} diagram}:
We can represent the \gls{abk:dpi} shown in \cref{fig:dpgeneralembodied} in the \emph{canonical form}, as shown in~\cref{fig:dpcanonical}. 
The meaning of the canonical form is captured by the categorical trace operator~$\mathsf{dp}= \trace{\mathsf{dp}'}$,
where~$\mathsf{dp}$ is the co-design problem with loops and~$\mathsf{dp}'$ is the co-design problem with no loops. This provides an upper bound for the complexity of the solution algorithm, as shown in~\cite[Proposition 5]{censi2015}. The solver (working principle of which is explained in detail in \cite{censi2015}) will solve the problem by creating a chain in the \gls{abk:poset} of antichains of $\bar{\mathbb{R}}_{\geq 0}^5$, where the latter represents the product \gls{abk:poset} containing the five resources which are fed back.
\paragraph*{Discussion}
We note three points. 
First, the approach described in this section is applicable to a large variety of systems in engineering, and describes a natural way to incorporate trade-offs in design decisions. Second, the modular and compositional properties of this tool allow different designers to focus on specific blocks, and to then intuitively interconnect the \gls{abk:dpi}. Finally, no specific mention of uncertainty was made, but the nature of the proposed framework allows for an extension in this direction, as explained in \cite{censi2017}.
\section{Co-design Results}
\label{sec:case_study}
In this section we showcase the abilities of the proposed framework to solve the co-design problem of an \gls{abk:av}. By considering the \gls{abk:dpi} proposed in~\cref{fig:dpgeneralembodied}, we optimize the design of an \gls{abk:av} by means of \Rtext{cost}, \Rtext{danger}, \Rtext{discomfort}, \Rtext{emissions}, robustness to \Ftext{environment} and \Ftext{cruise speed}. In particular, the design options, listed in~\cref{tab:designvars}, include the selection of longitudinal and lateral sensors, control parameters, vehicles, computers and detection algorithms. Note that the co-design optimization performed in this work is solved by a framework based on a formal language\footnote{Visit \url{co-design.science} for further information and future code release. The solution techniques and their complexity are described in~\cite{censi2015} and in the talk at \url{https://bit.ly/3ellO6f}.}.  
\begin{remark}
The following case study does not need to be analyzed quantitatively but qualitatively: our contribution lies  in the formalization of the meta-models and is not about numbers. We look forward to working with experts in each specific field to get more accurate data.  Also, the complexity for this kind of optimization problems is described in~\cite[Proposition 5]{censi2015}.
\end{remark}
\begin{figure}[t]
\begin{subfigure}[b]{\columnwidth}
\includegraphics[width=\linewidth]{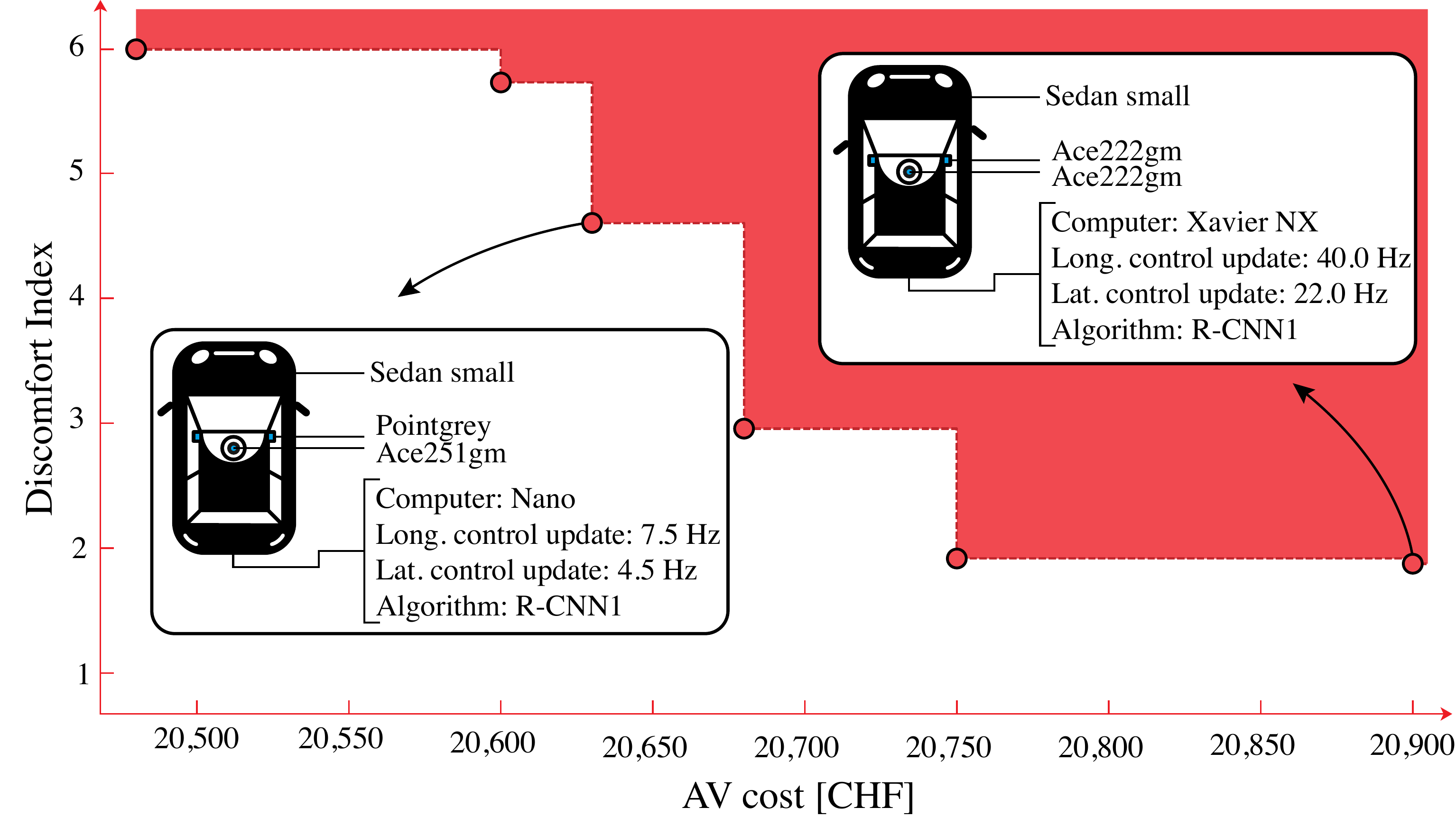}
\subcaption{Trade-off (antichain) of \Rtext{cost} and \Rtext{discomfort} in the design of an \gls{abk:av} able to drive during the day at \unit[55.0]{km/h}, with corresponding design choices.}
\label{fig:costdiscomfort}
\end{subfigure}
\begin{subfigure}[b]{\columnwidth}
\includegraphics[width=\linewidth]{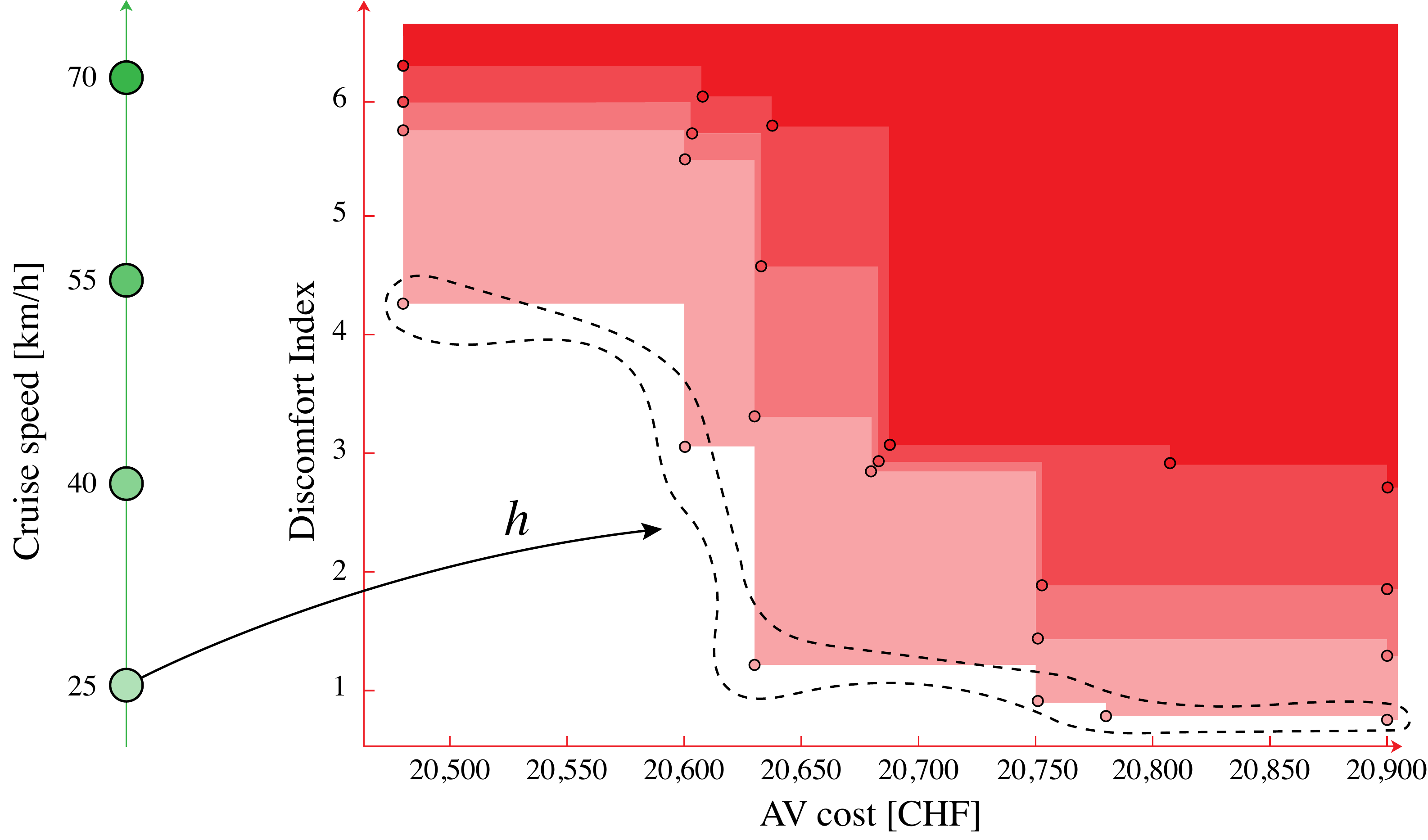}
\subcaption{Monotonicity of the \gls{abk:dpi}: Higher \Ftext{cruise speed} requires higher \Rtext{cost} and \Rtext{discomfort}.}
\label{fig:speedcostdisc}
\end{subfigure}
\caption{Trade-offs of \Ftext{cruise speed}, \Rtext{cost} and \Rtext{discomfort} in the design of an \gls{abk:av}. (a) The figure shows the antichain of optimal design solutions. The red dots represent the optimal design solutions and the colored area represents the upper sets of resources for which the \Ftext{cruise speed} of \unit[55.0]{km/h} is feasible. One can see selected highlighted implementations corresponding to specific points in the antichain. (b) Pareto fronts of \Rtext{resources} (expressed in terms of \Rtext{cost} and \Rtext{discomfort}) as a function of the provided \Ftext{cruise speed}. Monotonicity is expressed via inclusion of the drawn upper sets.}
\label{fig:results_first}
\vspace{-0.6cm}
\end{figure}

We assume obstacles to be distributed following a spatial Poisson process and generate sensor performance curves for the 16 sensors listed in~\cref{tab:designvars} (see \cref{fig:sensorslong} for an example) using the results from~\cite{braun2019,liu2019}. The authors of~\cite{braun2019} compare the pedestrian camera detection performance during day- and night-time at two different operating points of a Faster R-CNN object detection algorithm. The performance in terms of recall and precision is investigated against object distances and object heights in \unit[]{px}. We leveraged these results to interpolate the performance of the object detection with different sensors, using sensor resolution and angle of view~\cite{hartley2003}. For the camera accuracy we used the stereo depth error for a fixed baseline, assuming monocular camera depth estimation~\cite{gallup2008, zhou2017}. Lidar sensor performance curves are generated using the results of~\cite{liu2019}, who show that the 3D pedestrian detection efficiency decreases as the distance between lidar and objects increases. This effect can be explained by the decrease in number of reflected points available per obstacle as the distance increases. Recall and precision performances are presented for different distances for three different algorithms (KDE-based, STM-RBNN and STM-KDE). The authors trained and validated their models on the KITTI dataset~\cite{kitty2013}, originating from measurements of a Velodyne HDL-64 lidar. We estimated the measurement points per object at different distances and used them to interpolate for other lidars, generating curves similar to the ones presented in~\cite{borgmann2020}. Finally, we assumed a constant accuracy for the lidars, extracting it from sensor catalogues. Note that our framework can accommodate arbitrary sensor curves, enabling us to test non-existent sensor performances as well. Given this setting, we simulated both the longitudinal and lateral control of the \gls{abk:av}, creating numerical catalogues for the \gls{abk:dpi}. As previously explained, these simulations can be run in parallel, and build a database on which the solver can operate.
\paragraph*{Solutions which guarantee a certain speed (\cref{fig:costdiscomfort})}
We assume an obstacle density of \unit[5.0]{obstacles/km} during the day and query the optimal design solutions which enable the \gls{abk:av} to reach \unit[55.0]{km/h} (\cref{fig:costdiscomfort}). The dashed line represents the antichain of optimal solutions for the co-design problem, consisting of \Rtext{cost} and \Rtext{discomfort}. These solutions are not comparable, meaning that there is no instance which yields simultaneously lower \Rtext{discomfort} and \Rtext{cost}. In solid red we represent the upper set of resources. Attached, one can find selected design implementations, corresponding to specific optimal solutions. In general, as the budget for the \gls{abk:av} increases, one is able to reduce the discomfort. For instance, with a cost of of \unit[20,630]{CHF} one can obtain a discomfort index of 4.49, buying a small Sedan, equipping it with Pointgrey lateral cameras and an Ace251gm longitudinal camera paired with the R-CNN1 detection algorithm, together with a Nano computer. The vehicle is controlled at \unit[7.5]{Hz} longitudinally and at \unit[4.5]{Hz} laterally. 
Notably, investing only \unit[150]{CHF} more per \gls{abk:av} improves the discomfort by \unit[250]{\%}, requiring a Xavier NX computing unit, which controls the \gls{abk:av} at \unit[40.0]{Hz} and \unit[22]{Hz} longitudinally and laterally, respectively, using measurements of two Ace222gm cameras.
\paragraph*{Monotonicity of the \gls{abk:av} \gls{abk:dpi} (\cref{fig:speedcostdisc})}
We consider increasing \Ftext{cruise speeds} and analyze the evolution in trade-offs in \Rtext{cost} and \Rtext{discomfort}. As can be gathered from~\cref{fig:speedcostdisc}, we query the co-design solver for multiple performances, given arbitrary environments (here \unit[5.0]{obstacles/km} during the day). Specifically, for each \Ftext{functionality}, we compute the map $h$ which maps to the minimum antichain of resources which provide it (\cref{def:map_h}). Note that by increasing the desired \Ftext{cruise speed}, one increases the required resources, as can be observed from the dominating upper sets in increasing red tonality. Given such trade-offs and the discrete nature of our tool, one can reason about the \gls{abk:dpi} by distinguishing the desired objectives.

\begin{table}[h]
\begin{footnotesize}
\begin{center}
\begin{tabular}{lll}
\textbf{Variable} & \textbf{Options}& \textbf{Source}\\
Vehicle& Sedan Small, Large, BEV, SUV, Minivan & \cite{american2018}\\
Computers& Xavier, AGX, Nano, XavierNX&\cite{nvidia}\\
Control update& Brake: \unit[1.0-50.0]{Hz}, Lane: \unit[0.1-100.0]{Hz}&-\\
Sensors&\\
Lidars &Puck, HDL32E/64E&\cite{Velodyne}\\
&OS032/64/128, OS232/64/128&\cite{Ouster}\\
Cameras &Ace251gm/222gm/13gm/5gm/15um & \cite{basler}\\
& Flir Pointgrey&\cite{flir}\\
Algorithms & Cameras: R-CNN1, R-CNN3&\cite{braun2019,borgmann2020} \\
&Lidars: KDE, STM-RBNN, STM-KDE&\cite{liu2019}
\end{tabular}
\end{center}
\caption{Variables, options and sources for the \gls{abk:av} co-design problem.}
\label{tab:designvars}
\end{footnotesize}
\end{table}
\section{Conclusions}
\label{sec:conclusion}
This article leveraged a monotone theory of co-design to formulate and solve the optimal co-design of embodied intelligence platforms. We have shown how we can employ principles of functional decomposition to formalize embodied intelligence tasks as interconnected co-design problems. This work shows for the first time the ability of this monotone co-design theory to describe design problems arising from analytical relations, simulations and catalogues. We demonstrate the potential of our approach by modeling an \gls{abk:av}, composed of heterogeneous components, usually modeled using different, compartmentalized techniques.
\paragraph*{Outlook}
The methods we propose are intuitive to understand and use, and promise to unite different engineering disciplines under a common goal. Our vision is the one of compositional engineering, which would allow one to describe and co-design systems from signals~\cite{zardini2020compositional} all the way up to control systems~\cite{zardiniLCSS2020} and mobility networks~\cite{ZardiniLanzettiEtAl2020b,ZardiniEtAlBis2020,zardini2021analysis}. In particular, we are interested in leveraging the complexity properties described in~\cite{censi2015}, showcasing the compositional properties of our framework, by interconnecting the co-design problem of an \gls{abk:av} with the co-design problem of a urban mobility system. There, the \gls{abk:dpi} reported in \cref{fig:dpcanonical} would become part of a larger diagram of interconnected \glspl{abk:dpi}, and one could get design strategies at a further abstraction level.

\section*{Acknowledgments}
The authors would like to thank Alessandro Zanardi for the fruitful discussions.
\bibliographystyle{IEEEtran}
\bibliography{paper}
\clearpage
}
{

\section{Conclusions}
\label{sec:conclusion}
This article leveraged a monotone theory of co-design to formulate and solve the optimal co-design of embodied intelligence platforms. We have shown how we can employ principles of functional decomposition to formalize embodied intelligence tasks as interconnected co-design problems. This work shows for the first time the ability of this monotone co-design theory to describe design problems arising from analytical relations, simulations and catalogues. We demonstrate the potential of our approach by modeling an \gls{abk:av}, composed of heterogeneous components, usually modeled using different, compartmentalized techniques.
\paragraph*{Outlook}
The methods we propose are intuitive to understand and use, and promise to unite different engineering disciplines under a common goal. Our vision is the one of compositional engineering, which would allow one to describe and co-design systems from signals~\cite{zardini2020compositional} all the way up to control systems~\cite{zardiniLCSS2020} and mobility networks~\cite{ZardiniLanzettiEtAl2020b,ZardiniEtAlBis2020,zardini2021analysis}. In particular, we are interested in leveraging the complexity properties described in~\cite{censi2015}, showcasing the compositional properties of our framework, by interconnecting the co-design problem of an \gls{abk:av} with the co-design problem of a urban mobility system. There, the \gls{abk:dpi} reported in \cref{fig:dpcanonical} would become part of a larger diagram of interconnected \glspl{abk:dpi}, and one could get design strategies at a further abstraction level.

\bibliographystyle{IEEEtran}
\bibliography{paper}
}
\end{document}